\documentclass[final,1p,times]{elsarticle}

\usepackage{amssymb}
\usepackage{amsmath}

\usepackage[normalem]{ulem}
\useunder{\uline}{\ul}{}
\usepackage{pdflscape}

\usepackage{longtable}
\usepackage{url}
\usepackage{todonotes}
\usepackage[hidelinks]{hyperref}

\journal{Dentistry Review}

\begin{document}

\begin{frontmatter}



\title{
Generative artificial intelligence in dentistry: Current approaches and future challenges
}


\author[dcc,imfd]{Fabián Villena}
\affiliation[dcc]{organization={Department of Computer Science, University of Chile},city={Santiago},country={Chile}}
\affiliation[imfd]{organization={Millenium Institute Foundational Research on Data},city={Santiago},country={Chile}}

\author[uc,uandes]{Claudia Véliz}
\affiliation[uc]{organization={Dentistry School, Faculty of Medicine, Pontificia Universidad Católica de Chile},city={Santiago},country={Chile}}
\affiliation[uandes]{organization={Facultad de Odontología, Universidad de los Andes},city={Santiago},country={Chile}}

\author[uc]{Rosario García-Huidobro}

\author[uc,iibm]{Sebastián Aguayo \corref{cor1}}
\affiliation[iibm]{organization={Institute for Biological and Medical Engineering, Schools of Engineering, Medicine and Biological Sciences, Pontificia Universidad Católica de Chile},city={Santiago},country={Chile}}

\cortext[cor1]{Corresponding author}

\begin{abstract}
Artificial intelligence (AI) has become a commodity for people because of the advent of generative AI (GenAI) models that bridge the usability gap of AI by providing a natural language interface to interact with complex models. These GenAI models range from text generation - such as two-way chat systems - to the generation of image or video from textual descriptions input by a user.
These advancements in AI have impacted Dentistry in multiple aspects. In dental education, the student now has the opportunity to solve a plethora of questions by only prompting a GenAI model and have the answer in a matter of seconds. GenAI models can help us deliver better patient healthcare by helping practitioners gather knowledge quickly and efficiently. Finally, GenAI can also be used in dental research, where the applications range from new drug discovery to assistance in academic writing.  
In this review, we first define GenAI models and describe their multiple generation modalities; then, we explain and discuss their current and potential applications in Dentistry; and finally, we describe the challenges these new technologies impose in our area.
\end{abstract}




\end{frontmatter}



\section{Introduction}
\label{sec:introduction}

The advent of generative artificial intelligence (GenAI) models has revolutionized the way we interact with complex models, making it easier for people to solve knowledge-intensive tasks by simply conversing with intelligent systems \cite{cao2023comprehensive}. This shift towards a more human-like interface has led to widespread use of GenAI in various industries - including healthcare - where its incorporation has been applied in multiple teaching and practical settings \cite{Zhang2023}.

AI is the simulation of intelligence exhibited by humans that is processed by machines and can be utilized via multiple implementations \cite{mccarthy2007artificial}. AI is not a new technology and has been used, specifically in healthcare, since the 1950s \cite{Ghaffari2024}. However, it has mainly been employed in a \textit{behind the scenes} fashion, without direct interaction with patients or practitioners. While often associated with complex mathematical functions, AI can also be implemented by simple human-defined rules that simulate processes. However, most people's understanding of AI is tied to machine learning (ML), a subset of AI that enables software to automatically generate mathematical functions through algorithms, mimicking human-like decision-making without explicit rules \cite{Shinde2018}. Today, most remarkable AI systems rely on ML to achieve impressive performance. Contemporary AI is perceived as a conversational technology that enables humans to engage in straightforward discussions and receive relevant responses \cite{cao2023comprehensive}. This shift towards more accessible AI has led to widespread usage over the past two years. Gone are the days when AI models were relegated to \textit{behind the scenes} operations; now, these complex softwares can be openly shared with the public, allowing anyone to interact with them through a user-friendly interface.

The evolution of AI began with symbolic systems, which relied on human-readable representations of problems (i.e., rule-based methods) to achieve a desired goal. While this approach provided excellent explainability, it faced scalability challenges when dealing with complex processes and required extensive human expertise to craft many rules. In a next phase, ML emerged, which enabled algorithms to learn these rules automatically from training datasets and generate models that produce desired outputs from input features. Most recently, the generative paradigm has gained prominence, where AI models are trained not to mimic specific processes but to comprehend human instructions and provide answers that satisfy those instructions.

In this context, AI and GenAI have been successfully applied to various aspects of dentistry for many years, with applications ranging from predictive models that analyze patient features and identify treatment outcomes \cite{Kurniawan2021} to caries detection models using bitewing images \cite{Khnisch2021}. Furthermore, generative approaches have emerged as a powerful tool for tackling a wide range of educational, administrative, research, and clinical challenges in the dental domain, which is the primary focus of this review. Particularly, the healthcare field presents unique ethical and privacy considerations that require heightened attention when exploring the widespread use of GenAI. As a result, implementing GenAI models in dental care comes with distinct challenges not typically encountered in other domains \cite{Kumar2024}. Ensuring that the deployed models protect patient health and privacy is essential, highlighting the need for careful consideration and oversight in these regards. Therefore, this review will also focus on the critical challenges the field must address to integrate GenAI into the dental domain successfully.

Therefore, this review will initially give a brief introduction regarding the theoretical definitions of AI and the GenAI paradigm, considering their multiple modalities. Next, we introduce and discuss the applications of GenAI in multiple dental fields ranging through dental education and student training, diagnosis and treatment planning, patient communication, and dental research. Finally, we describe some of the current and future challenges associated to the application of GenAI in the oral and dental field.

\section{Artificial intelligence}
\label{sec:ai}

AI is a subfield of computer science that aims to develop intelligent systems capable of performing tasks that typically require human cognition such as learning, reasoning, and problem-solving. AI seeks to replicate real-world processes governed by complex mathematical functions, such as $f(x)$, to determine a patient's disease prognosis based on multiple features. These functions can be arbitrarily complex, returning values that may include classes (e.g., ``sick" or ``not sick"), numbers, or even more elaborate natural language descriptions.

In AI, we approximate these complex functions, denoted as $\hat{f}(x)$ using parametric models, such as a simple linear regression $\hat{f}(x) = ax + b$, where we search the $a$ and $b$ parameters that minimize errors. Other elaborate parametric functions can better approximate the $f(x)$ target function. Neural networks, for example, are particularly effective at approximating complex functions such as describing patient images. On the other hand, ML occurs when an algorithm automatically adjusts the parametric function's parameters using process observations. When explicitly using neural networks, we refer to this process as deep learning (DL). For a more comprehensive and in-depth overview of AI, please consult a previously published review \cite{Ghaffari2024}.

\begin{figure}[]
\centering
\includegraphics[width=0.8\textwidth]{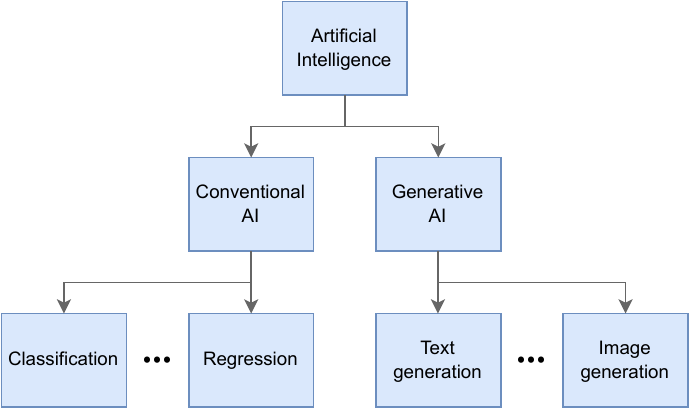}
\caption{Taxonomy of AI categories.}\label{fig:ai}
\end{figure}

We can categorize AI into two primary categories based on the nature of the output of $f(x)$ or our approximation $\hat{f}(x)$: \textit{conventional AI} and \textit{GenAI} (see Figure \ref{fig:ai}). Conventional AI involves predicting structured values from predefined categories, such as classifying patients as ``sick" or ``not sick" (also referred to as classification), or predicting numerical values such as the suggested number of months until a patient's next follow-up appointment (also referred to as regression). In contrast, GenAI generates unstructured data, including natural language sentences, images, videos, and audio \cite{cao2023comprehensive}, and is the specific type of AI that will be covered and discussed throughout the present review.

\section{Generative artificial intelligence (GenAI)}
\label{sec:genai}

\begin{figure}[]
\centering
\includegraphics[width=0.75\textwidth]{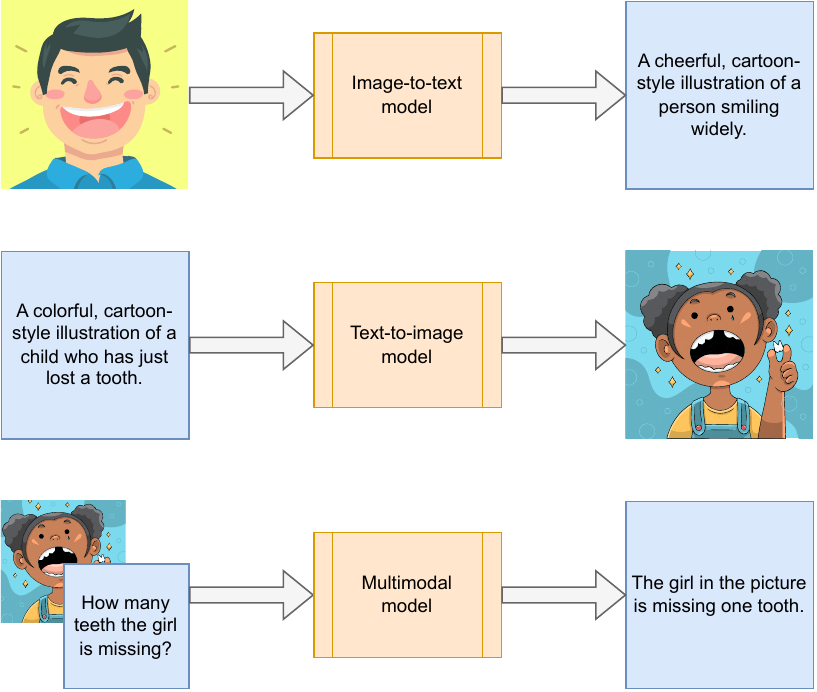}
\caption{Examples of GenAI models.}\label{fig:genai}
\end{figure}

GenAI models produce unstructured data like text, images, or audio. This generation process often involves satisfying a natural language instruction or description given by a human, where the model is prompted to create content that meets specific user-defined criteria. Alternatively, GenAI models can also mimic the style and characteristics of the data they were trained on without explicit guidance or input \cite{cao2023comprehensive, zhang2023complete}. See Figure \ref{fig:genai} for examples.

The concept of GenAI models has a rich history in computer science, with roots tracing back to the 1950s. However, it was only with the emergence of DL that GenAI models experienced a significant breakthrough. The flexibility of DL techniques enabled them to effectively approximate complex functions that underlie real-world phenomena, leading to substantial improvements in performance. Furthermore, the widespread availability of large datasets and extensive computing resources has fueled another wave of progress in DL-based models, particularly those focused on data generation. Some of the most relevant foundation approaches for content generation are described in the following paragraphs. All of these approaches are DL-based.

\paragraph{Generative adversarial networks (GANs)}
These approaches were among the first to show high-quality content generation, particularly in image generation. GANs consist of two parts: a generator and a discriminator. The generator is a neural network that attempts to learn the distribution of training examples in order to generate new data, and the discriminator determines whether the input is real or not. The generator tries to generate data that the discriminator considers as real \cite{goodfellow2014}.

\begin{figure}[]
\centering
\includegraphics[width=0.8\textwidth]{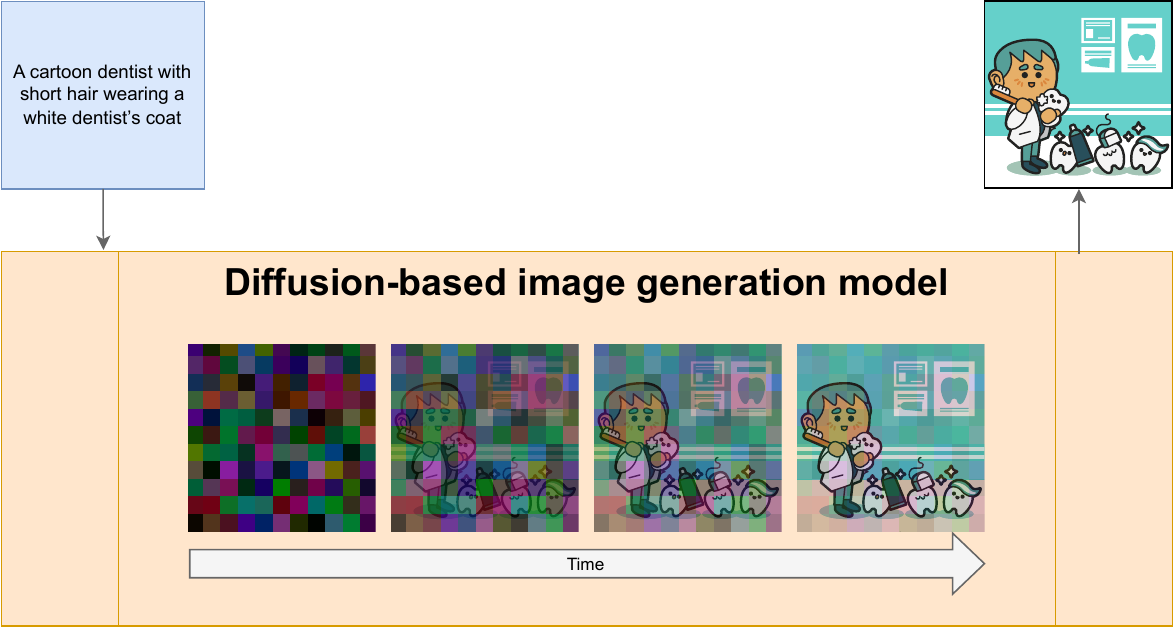}
\caption{Description of the image generation process through diffusion.}\label{fig:diffusion}
\end{figure}

\paragraph{Generative diffusion models (Diffusion)}
Diffusion-based approaches have gained popularity primarily due to their application in image generation tasks. The underlying mechanism of these models is based on an innovative concept: the model, for training, intentionally corrupts the training data by incrementally adding noise perturbations and then learns to reverse this process (see Figure \ref{fig:diffusion}). By doing so, the model discovers a way to progressively refine the noisy data until it converges to a realistic image, effectively generating new data that resembles the original \cite{ho2020}.

\paragraph{Transformer}
This type of neural network architecture is mainly used for text encoding and generation (also called decoding). It consists of an encoder that takes an input sentence and produces a hidden numerical representation. A decoder then uses this representation to generate a new output sentence. One of the key features of the transformer is its self-attention mechanism, which allows the model to prioritize important words in the input sentence for better representation of meaning in the hidden numerical form. The representations generated by the encoder are helpful for text generation and can also be applied to a wide range of other tasks involving information extraction from text. Additionally, these representations can be used to create images with diffusion models based on textual descriptions \cite{vaswani2017}.

\subsection{Text generation}
\label{sec:text}

\begin{figure}[]
\centering
\includegraphics[width=0.8\textwidth]{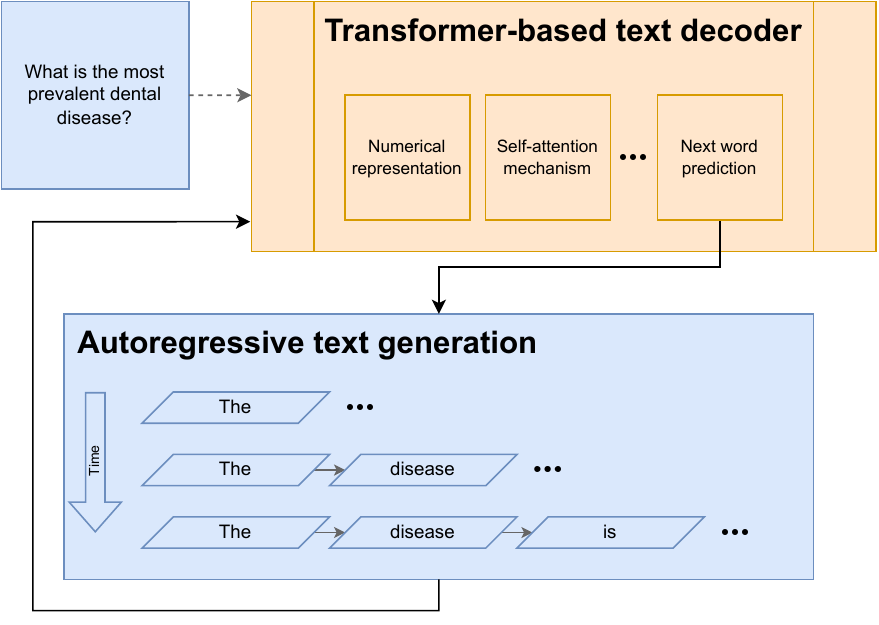}
\caption{Description of the transformer architecture and the autorregressive text generation mechanism.}\label{fig:transformer}
\end{figure}

Generating text content relies on causal language models that utilize a given textual context to predict the most probable next word \cite{Li2024}. This output is then fed back into the model to predict another subsequent word, a technique known as autoregression (see Figure \ref{fig:transformer}). At the core of these autoregressive causal language models lie decoders based on the transformer architecture, which employs self-attention mechanisms \cite{vaswani2017}.

The size of a model is typically measured by the number of internal parameters within its neural network. Interestingly, research has shown that increasing the size of causal language models leads to improved performance \cite{kaplan2020scaling}. Moreover, large language models exhibit emergent abilities, which are capabilities not present in smaller models but arise when scaling up \cite{wei2022emergent}. These emergent abilities include in-context learning, where models can generate expected outputs based on natural language instructions without additional training; instruction following, where models perform well on unseen tasks described through instructions; and step-by-step reasoning, where models can solve complex problems by breaking them down into intermediate reasoning steps.

Text generation models are not limited to processing textual information alone. They can also incorporate data from other modalities, such as images, to fulfil instructions based on both textual context and visual inputs. This capability enables the models to generate more comprehensive and accurate responses that consider multiple sources of information.

\subsection{Image generation}
\label{sec:imagevideo}

While Generative Adversarial Networks (GANs) \cite{goodfellow2014} and diffusion-based models \cite{ho2020} have primarily been employed for generating images and videos (sequence of images), their approach to controlling the creative process differs. GAN-based generators typically take some not-so-flexible input to condition the generation process, allowing for some control over the output. In contrast, diffusion-based generators offer a higher degree of fine-grained control over the image-generation process, enabling the incorporation of specific inputs and cues to shape the output in the form of natural language descriptions. This distinction enables diffusion-based models to produce more tailored and nuanced results, whereas GANs may rely on chance and their predefined training data to drive the creative process.

Diffusion-based image generators use transformer-based text encoders to represent the input natural language description and then decode this representation as an image through diffusion. Besides receiving only textual descriptions as input for image generation, diffusion-based models can also receive images and text to guide image generation.

\subsection{Graph generation}
\label{sec:othermodalities}

GenAI techniques can generate other forms of data, including audio, music, and graphs \cite{cao2023comprehensive}. However, the generation of graphs is of significant importance in health science. Graphs are a data structure representing information as a network of interconnected vertices and edges, where each vertex can be any object or entity, and an edge represents the relationship between them. This data modality is particularly valuable in health science, as it allows for the representation of complex relationships between different entities, such as atoms in a molecule and processes in a metabolic pathway.

Graph generation has the potential to revolutionize drug discovery by seamlessly integrating highly structured molecule graphs with natural language descriptions, enabling researchers to comprehend molecular knowledge better and efficiently explore molecule structures. By leveraging transformer-based text encoders, molecule generation from text descriptions can be achieved through a process in which transformer-encoded textual information is fed into a decoder to generate corresponding graphs that accurately represent molecules \cite{zhang2023graph}.

\section{Current use of GenAI in dental education}
\label{education}

\begin{figure}[!h]
\centering
\includegraphics[width=0.8\textwidth]{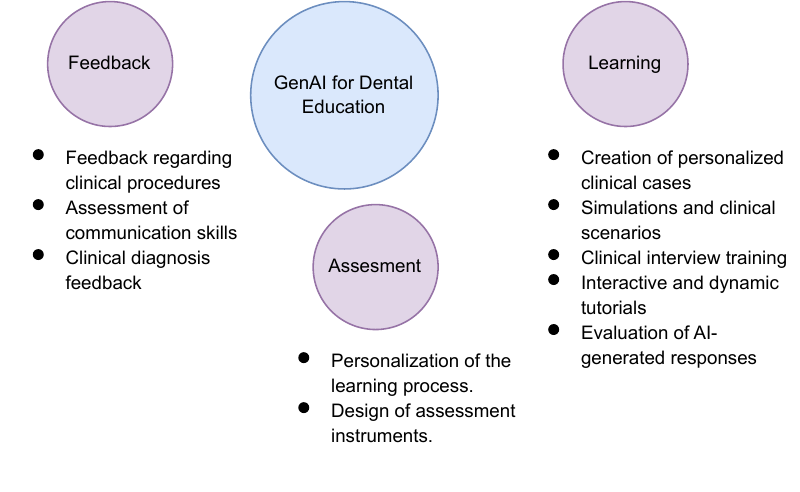}
\caption{Summary of the applications of GenAI in Dental Education.}\label{fig:dental_education}
\end{figure}

In recent years, GenAI has rapidly transformed both content generation and teaching implementation in the field of dental education \cite{Gordon2024}, providing innovative tools to personalize and enrich the teaching-learning and assessment processes. These technologies allow educational content to be adapted to the individual needs of each student, thus improving the efficiency and effectiveness of training in dentistry and health sciences in a more personalized manner \cite{Thorat2024}. The new strategies implemented through GenAI range from creating detailed simulations that facilitate pre-clinical practice, to developing systems that provide an immediate feedback on clinical skills following dental treatment. However, as the case with other highly disruptive technologies, the implementation of these GenAI approaches is not without challenges. Resistance to change, defining roles and responsibilities, ethical concerns about privacy and academic integrity, and technological dependency are just some of the obstacles that must be overcome to effectively integrate GenI into the dentistry curriculum in an effective manner. Thus, this section will describe some of the currently used approaches that utilize GenAI to support and strengthen dental education. Figure \ref{fig:dental_education} and Table \ref{tab:education} contain a summary of the applications described in this section.  

\subsection{Assesment}

The various applications of GenAI can aid in designing appropriate instruments and activities that correctly evaluate theoretical and clinical skills obtained by dental students during the course of their studies, effectively considering learning content and outcomes \cite{Gordon2024, Bannister2023, Uribe2024, Saghiri2021, Kim2023, Ali2023}. Some relevant examples include: 


\begin{description}
    \item[Personalization of the learning process] GenAI can develop questions that fit each student's learning progress in a personalized manner. In dentistry, where technical and practical knowledge is essential, AI can adjust the difficulty level of questions based on the student’s previous performance \cite{Cheung2023,Simms2024}. This way, each student can improve on their weaknesses and limitations in a student-specific targeted approach.
    \item[Design of assessment instruments] GenAI can create open-ended or structured response questions that assess different learning outcomes at various levels of Bloom's taxonomy \cite{krathwohl2002revision}, along with guidelines for grading these instruments. It can also develop detailed and adaptive rubrics and guidelines for specific tasks, assessing both theoretical knowledge and the specific clinical skills required for dental practice \cite{Indran2023,Cheung2023}. This does not only aid in generating effective examination strategies, but also optimizes the time needed to create and implement these instruments.
\end{description}

\subsection{Feedback}

Besides the assessment of learning outcomes, one of the most current widespread uses of GenAI is providing immediate feedback to students based on their performance \cite{Chheang2024,Agarwal2023,Parker2023} and clinical skills \cite{Gordon2024, Thorat2024, Bannister2023, Kim2023}. 

\begin{description}
    \item[Feedback regarding clinical procedures] By using GenAI, the performance of students following the pre-clinical simulation of dental therapeutic procedures can be analyzed. Among other parameters, GenAI can evaluate technique qulality, accuracy, and adherence to clinical protocols, and provide detailed and personalized feedback on each aspect of the procedure.
    \item[Assessment of communication skills] GenAI can also be prompted to simulate interactions with patients, evaluating the communication abilities of students. For this, GenAI can analyze the clarity of information provided, and the ability to empathize with the simultated patient, therefore training dental students in the correct way of interacting during the clinical practice setting.
    \item[Clinical diagnosis feedback] Finally, GenAI can assess student ability of reaching accurate diagnoses based on simulated clinical data, providing specific feedback on the reasoning and logic used to reach that particular diagnosis. These tools can be particularly useful for disciplines such as oral pathology and oncology, as well as support the effective treatment planning for restorative approaches.
\end{description}

\subsection{Learning}

On the other hand, GenAI can also play a crucial role during the learning experience of dental students. For example, some systems can adapt learning resources to the individual needs of students, which is important in scenarios where learning requirements can vary widely among individuals \cite{Gordon2024, Thorat2024, Bannister2023, Tricio2024, Uribe2024, Saghiri2021, Mu2024}. With these tools it is possible to promote self-directed learning, by providing materials and resources that adapt to the individual needs of each student \cite{Gordon2024, Thorat2024, Preiksaitis2023}. Additionally, GenAI can also encourage the development of critical thinking during dental training by providing an interactive environment that challenges students to assess and verify the information \cite{Thorat2024, Uribe2024, Saghiri2021}. Most importantly, GenAI can be used to create specific simulations that allow students to practice clinical procedures in a controlled and secure pre-clinical environment \cite{Bannister2023, Preiksaitis2023, Kim2023, Mahrous2023} to develop abilities and skills related to the clinical interview process, critical thinking, and differential diagnosis, among others \cite{Thorat2024, Uribe2024, Mu2024, Borra2007, Schropp2023}. 

\begin{description}
    \item[Creation of personalized clinical cases] Using GenAI, it is possible to create realistic simulations of clinical cases that can be adjusted to the skills and comprehension level of every student. For example, for an early-career student, GenAI could generate simpler cases that include step-by-step feedback. For more advanced students and postgraduate students, generated cases can increase in complexity and include fewer guides in order to develop higher skills according to their knowledge levels \cite{Lim2023,Yanagita2024}.
    \item[Simulations and clinical scenarios] Through the creation of realistic clinical simulations with GenAI, students are able to confront complex situations that require them to apply their knowledge to make the correct diagnostic and treatment decisions. This not only improves their technical skills but also enhances their ability for critical thinking.  
    \item[Clinical interview training] By using GenAI models, virtual patients with diverse and complex medical records can be created. These simulated patients can even present with realistic symptoms and respond to questions in a manner similar to a real patient, enabling students to practice medical record keeping and effective communication in a protected and controlled pre-clinical environment \cite{Hanai2024,Surez2022}. This also reduces the need to use real patient data, helping support data confidenciality and protection guidelines during clinical training of dental students.
    \item[Interactive and dynamic tutorials] Contrary to conventional assessments, tutorials developed by GenAI can be designed to adapt in real time to student answers for a more personalized approach. For example, if a student shows difficulties with a particular concept (such as a particular dental anesthesia technique), GenAI can offer additional explanations or modify the content to reinforce the necessary knowledge before moving forward \cite{Fang2024,Kavadella2024}. 
    \item[Evaluation of AI-generated responses] Strategies that incorporate responses from a GenAI-based chatbot to urge students to verify information with reliable sources, to assess the reliability of the model, challenging students to identify errors or inaccuracies in these responses promoting critical analysis skills and detailed evaluation of information.  
\end{description}

\begingroup\small
\begin{longtable}{cp{3.0cm}p{2.5cm}p{2.5cm}p{1.5cm}}
\hline
\textbf{Citation} & \textbf{Goal} & \textbf{Input} & \textbf{Output} & \textbf{Model} \\ \hline \endhead
{\ul Assessment} &  &  &  &  \\
 \cite{Indran2023} & Generate diverse and high-quality medical questions. & Description of the type questions and its core concepts or clinincal scenarios. & Questions. & GPT-4.0 \\
 \cite{Cheung2023} & Assess the quality of multiple-choice questions (MCQs) created by GenAI & Instruction to write multiple-choice question based on some criteria and references. & Multiple-choice questions. & ChatGPT \\
 \cite{Simms2024} & Integrate AI tools, into prelicensure nursing education & Prompts from students for various activities such as questions and nursing care scenarios. & Educational content. & ChatGPT Plus \\ 
{\ul Feedback} &  &  &  &  \\
 \cite{Chheang2024} & Enhance human anatomy education by introducing GenAI-based embodied virtual assistants. & Questions in verbal form related to human anatomy in various complexity levels. & Detailed answers to the anatomy-related questions and explanations about anatomical structures. & GPT-3.5 \\
 \cite{Agarwal2023} & Evaluate the proficiency of GenAI models in answering and explaining multiple-choice questions (MCQs) in physiology. & A set of 55 MCQs from 10 competencies of medical physiology. & Answers and detailed explanations for the 55 MCQs. & GPT-3.5 and Claude-2 \\
 \cite{Parker2023} & Explore the suitability of a GenAI model for Automated Writing Evaluation in scholarly writing. & Papers written by students and instructions for grading. & Scores and feedback for each paper. & GPT-3.0 \\
 {\ul Learning} &  &  &  &  \\
 \cite{Fang2024} & Evaluate the enhancement of the learning experience with the use of GenAI tools. & User queries related to clinical protocols and procedures. & Responses to the clinical questions posed by the students. & Custom-tailored chatbot. \\
 \cite{Lim2023} & Evaluate the capacity of GenAI models to produce realistic images pertinent to cosmetic surgery and to assess their potential as educational tools. & Instructions to generate images representing ideal standards for various cosmetic procedures. & Images of noses, faces, and eyelids embodying the pinnacle of cosmetic appeal. & DALL-E 2, Midjourney and Blue Willow \\
 \cite{Hanai2024} & Evaluate the potential of GenAI in facilitating clinical communication in addressing sensitive topics. & Instruction to generate questions based on data regarding sexual difficulties among cancer survivors. & Questions. & GPT-3.5 \\
 \cite{Kavadella2024} & Evaluate the implementation of GenAI in the educational process of undergraduate dental education. & Instructions related to assignment topics. & Detailed explanations, summaries and references regarding the queries topics. & ChatGPT \\
 \cite{Surez2022} & Evaluate the use of GenAI to simulate a virtual patient to enhance dental students' diagnostic skills. & Expressions and intents related to diagnosing pulp pathology. & Responses based on the provided inputs, simulating a conversation with a patient. & Dialogflow \\
 \cite{Yanagita2024} & Investigate whether a GenAI model can generate comprehensive illness scripts for multiple diseases. & Instructions to generate illness scripts for each disease. & Illness scripts adhering to the specified format. & GPT-4.0 \\ \hline
\caption{Studies exploring the application of GenAI in dental education.}\label{tab:education}
\end{longtable}
\endgroup

\section{Enhancing the dental clinical practice with GenAI}
\label{practice}

\begin{figure}[]
\centering
\includegraphics[width=0.8\textwidth]{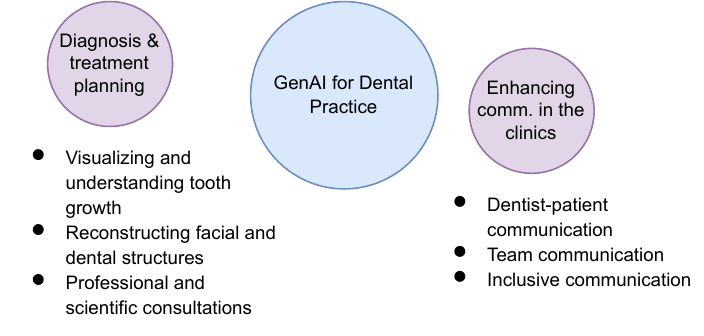}
\caption{Summary of the applications of GenAI in Dental Practice.}\label{fig:dental_practice}
\end{figure}

Additionally to the previously discussed uses, GenAI has become a revolutionary tool in the field of healthcare, offering advanced solutions to improve clinical outcomes and efficiency in patient care.
In this context, GenAI can provide real-time assistance during clinical procedures, guiding healthcare professionals with recommendations based on a broad set of clinical data and scientific evidence. This can not only improve the accuracy and safety of procedures but also reduce the risk of human error. The various applications of GenAI in clinical dentistry will be reviewed, considering its contributions to diagnosis, treatment planning and communication \cite{Moulaei2024}. It is important however to consider that, despite these advances, most of the available literature emphasizes the need of further research to solidify the clinical use of GenAI in a reproducible and standardizable manner. Figure \ref{fig:dental_practice} and Table \ref{tab:practice} contain a summary of the applications described in this section. 

\subsection{Diagnosis and treatment planning} 
GenAI technologies are revolutionizing dentistry by significantly enhancing diagnostic accuracy and treatment planning in a variety of clinical disciplines. These tools enable dental professionals to clearly visualize the overall condition of patients, leading to more precise diagnoses and tailored treatment plans. For example, GenAI-driven imaging has been used to predict dental development in children \cite{Kokomoto2024} and to assist in the reconstruction of facial structures \cite{Arjmand2023}, thereby improving clinical outcomes. Additionally, AI technology can aid the clinician by offering real-time, evidence-based consultations, optimizing both aesthetic and functional results \cite{Makrygiannakis2024,Giannakopoulos2023}. By integrating GenAI into the clinical practice, dentistry is expected to continue advancing in treatment precision, effectiveness, and patient engagement, ultimately elevating standard of care and patient satisfaction.

\begin{description}
    \item[Visualizing and understanding tooth growth] GANs have been utilized to predict dental development in children from panoramic radiographs. These networks can be used to generate synthetic images illustrating the transition from primary to permanent dentition, and provide clinicians with a valuable tool for better visualizing and understanding tooth growth, enhancing treatment planning, and improving patient education in pediatric dentistry \cite{Kokomoto2024}.
    \item[Reconstructing facial and dental structures] Replacing damaged or lost tissues remains one of the greatest challanges in regenerative medicine and dentistry. Thus, the use of deep neural networks can enable the prediction of the three-dimensional shape of the face in cases of complex facial deformities caused by trauma, cancer, infections, or congenital anomalies \cite{Arjmand2023}. Similarly, GenAI has shown feasibility as a novel approach to design single-molar dental prostheses to replace missing teeth in a patient-specific approach \cite{broll2024data,chau2024accuracy}.
    \item[Professional and scientific consultations] Finally, the advent of GenAI chatbots represents a promising development for dental professionals, offering the potential to serve as ``chairside scientific consultants" in real time during clinical interventions. Some of these AI tools are being designed with the aim of generating evidence-based responses that can provide dentists with immediate access to relevant clinical information and recommendations during patient consultations \cite{Makrygiannakis2024,Giannakopoulos2023}. Nevertheless, these technologies are still in early phases of development, and require important improvements in order to make them more user-friendly, intuitive, and effective at providing scientifically-backed advice and guidance.
\end{description}

\subsection{Enhancing communication in the clinics}
Current GenAI approaches aim to enhance communication in dental practice, particularly benefiting patient interactions. GenAI can explain and illustrate complex treatment plans to patients by generating detailed visualizations and personalized simulations, increasing transparency and patient trust \cite{sai2024generative}. Additionally, GenAI has the potential to improve team communication by integrating complex data sets to create precise predictive models, helping dental teams align their goals and methods for cohesive and efficient teamwork. Furthermore, GenAI could serve as a tool for generating visual aids tailored to patients with language barriers or specific needs, potentially improving their understanding and engagement with their practitioner. Overall, the application of GenAI in these areas demonstrates its potential to revolutionize dental practice by fostering better understanding and collaboration among patients, clinicians, and supporting personnel.

\begin{description}
    \item[Dentist-patient communication] Recent advances in GenAI have began to promote dentist-patient communication by providing an array of tools for an enhanced understanding of treatment plans. 
    This facilitates informed decision-making and increases patient satisfaction through personalized simulations \cite{sai2024generative}. Additionally, Gen AI is used to generate detailed, patient-specific post-operative instructions that can promote individual compliance to treatment. This tool allows healthcare professionals to provide clear and personalized information, enhancing patient comprehension and facilitating post-surgical recovery. By tailoring instructions to individual needs, ChatGPT has been shown to ensure precise and accessible guidance for patients, optimizing clinical outcomes \cite{mohan2023utility}.
    \item[Team communication] Coordination among members of the clinical team could potentially be enhanced by utilizing models to predict treatment outcomes through the integration of complex data, such as computed tomography images and patient-specific characteristics. These detailed predictions might help dentists, lab technicians, and other medical professionals align their goals and methods, ensuring cohesive teamwork and optimizing both clinical results and team efficiency.
    \item[Inclusive communication] In addition to the aforementioned advantages, GenAI could potentially enhance communication for patients with specific needs or language barriers by creating bespoke tailored visual aids. These aids, such as multilingual annotated images and interactive models, could bridge communication gaps and ensure that patients fully understand their treatment plans and procedures. This inclusivity could foster greater patient engagement and compliance, ultimately leading to improved overall outcomes in dental care.
    
\end{description}

\begingroup\small
\begin{table}[!h]
\begin{tabular}{cp{3.0cm}p{2.5cm}p{2.5cm}p{2.0cm}}
\hline
\textbf{Citation} & \textbf{Goal} & \textbf{Input} & \textbf{Output} & \textbf{Model} \\ \hline
{\ul D\&T} &  &  &  &  \\
\cite{Kokomoto2024} & Predict dental growth using panoramic radiography. & Panoramic radiography. & Panoramic radiography at a further dental growth stage. & StyleGAN-XL \\
\cite{Arjmand2023} & Underlying bone geometry and additional metadata. & Panoramic radiography. & 3D face shape. & U-Net \\
\cite{chau2024accuracy} & Investigate the accuracy of a GenAI system in designing biomimetic single-molar dental prostheses. & 3D mesh of a cast without a first molar. & 3D mesh of a first molar. & GAN \\
\cite{broll2024data} & Data-driven approach for the partial reconstruction of occlusal surfaces. & 2D projection of a 3D tooth preparation. & Reconstructed 2D tooth representation. & StyleGAN-2 \\
{\ul Comm.} &  &  &  &  \\
\cite{Makrygiannakis2024} & Evaluate the potential of GenAI language models in orthodontics. & Open clinical questions related to orthodontics. & Responses to the clinical questions. & ChatGPT 3.5 \& 4, Google Bard, Bing Chat \\
\cite{Giannakopoulos2023} & Evaluate the performance of GenAI language models in supporting evidence-based dentistry. & Open clinical questions related to dentistry. & Responses to the questions. & ChatGPT 3.5 \& 4, Google Bard, Bing Chat \\
\cite{mohan2023utility} & Evaluate the performance of GenAI language models in generating post-operative instructions for facial trauma patients. & Prompt to generate instructions. & Post-op instructions for mandibular fracture, maxillary fracture, nasal fracture, facial laceration. & ChatGPT \\ \hline


\end{tabular}
\caption{Studies exploring the application of GenAI in dental practice.}\label{tab:practice}
\end{table}
\endgroup

\section{GenAI in dental and medical research}
\label{research}

Besides its uses in teaching, training, and clinical practice, GenAI is emerging as an interesting alternative to support dental and craniofacial research. More specifically, researchers are utilizing diverse GenAI tools to facilitate the development and execution of projects. GenAI is particularly promising to aid in the execution of time consuming tasks such as revising the literature and optimizing data analysis and processing. Table \ref{tab:research} contain a summary of the applications described in this section.

\subsection{Scientific writing}

 The current and potential support of GenAI in the dental research process spans from the review and synthesis of the literature, the development of research questions, the creation of datasets, the optimization of image analysis techniques, and the publication and dissemination of research results \cite{Morley2023} (Figure 1). Regarding scientific writing, platforms such as ChatGPT, Bard, and Bing Chat have been used to aid in the development of research questions, hypotheses, and bibliographic reviews in the context of medical healthcare \cite{vanDis2023}. The use of GenAI to support scientific writing is also expected to significantly support researchers whose primary language is not English \cite{Conroy2023, Babl2023, Hsu2023}, although authors are expected to disclose its use for transparency reasons \cite{Tang2023}. In this reagard, its use can also prove useful to train young students and early-career reserchers in the mechanics of scientific writing (e.g., grant proposals, manuscript preparation, case study presentations, etc).

\begin{figure}[]
\centering
\includegraphics[width=0.8\textwidth]{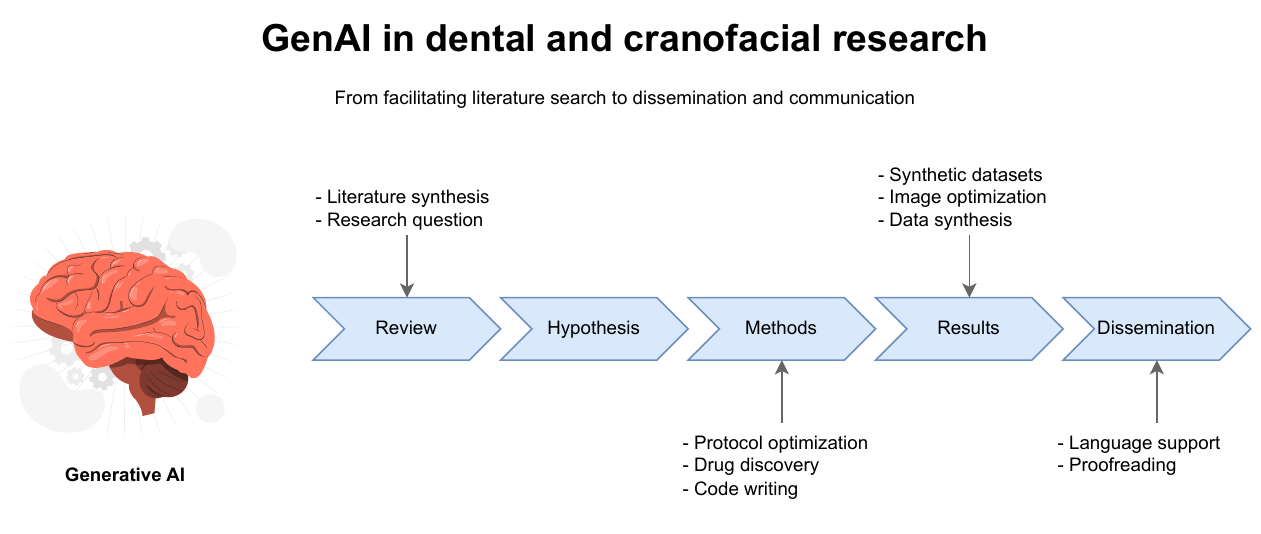}
\caption{Summary of the application of GenAI in dental and craniofacial research.}\label{fig:research}
\end{figure}

\subsection{Aiding the search for novel biomolecules and therapies}

On the other hand, GenAI can greatly facilitate the drug discovery pipeline, with a profound impact on the dental and medical fields. In this context, the use of GenAI for the development of novel molecules and the generation of large synthetic datasets are among the most explored approaches in the literature. For example, the use of GenAI - such as the Chroma or RoseTTAFold diffusion-based models – allows for the design and synthesis of \textit{de-novo} proteins for targeted drug discovery and personalized medicine applications \cite{Watson2023}. Also, the structure prediction of proteins from their amino acid sequences can currently be carried out with AlphaFold, a GenAI platform that has opened the door to understanding the structure-function relationship of molecules associated with healthy and diseased states \cite{Jumper2021}. The most recent version of Alphafold, Alphafold 3, can predict not only the structure but also the biological interactions between proteins, DNA, and RNA. The potential of these approaches for the discovery of novel molecules to treat common dental conditions such as orofacial pain, dentinal hypersensitivity, and biofilm-mediated diseases remains to be fully explored. On the other hand, the use of GenAI for generating large datasets of synthetic patients and conditions is supporting innovation in medical education research, particularly in the field of diagnostics \cite{Huang2023}. Finally, GenAI has also proven to be  highly valuable for the optimization of imaging-based research – including MRI, CT scans, and microscopy – in order to increase image accuracy and resolution \cite{PintoCoelho2023}. These advances are expected to open new avenues of translational research in the coming years, with a profound impact in reshaping clinical practice in the fields of oral radiology and diagnostics.

\begingroup\small
\begin{table}[!h]
\begin{tabular}{cp{3.0cm}p{3.0cm}p{2.0cm}p{1.5cm}}
\hline
\textbf{Citation} & \textbf{Goal} & \textbf{Input} & \textbf{Output} & \textbf{Model} \\ \hline 
{\ul Writing} &  &  &  &  \\
\cite{Babl2023} & Produce a quality conference abstract using a ﬁctitious but accurately calculated data table. & Prompt to write an abstract from a data table. & An abstract. & ChatGPT \\
\cite{Hsu2023} & Write a short paper. & Multi-stage prompts to write a short paper. & A short paper. & ChatGPT \\
{\ul Discovery} &  &  &  &  \\
\cite{Watson2023} & Design of protein structures and functions with diffusion-based models. & Simple molecular specifications. & Protein backbones. & RFDiffusion \\
\cite{Jumper2021} & Develop a computational approach capable of predicting protein structures to near experimental accuracy. & Primary amino acid sequence and aligned sequences of homologues. & Protein structures. & Transformer \\ \hline
\end{tabular}
\caption{Studies exploring the application of GenAI in dental research.}\label{tab:research}
\end{table}
\endgroup

\section{Challenges of GenAI in the dental field}

The current integration of GenAI into the biomedical field will undoubtedly revolutionize how dentistry is taught, practised, and researched. However, its implementation faces a series of challenges beyond the technology itself that must be addressed in the short and long term. From the resistance to change to ethical considerations and the need to promote AI literacy, educators, dental practitioners, and researchers are at a crucial point where they must address these aspects to ensure effective and ethical adoption of AI across educational, research and clinical settings.

\begin{description}
    \item[Resistance to change] This is a common challenge when AI is introduced into any field, often stemming from concerns that AI will displace traditional methods or undermine professional roles. For instance, academics may worry that AI threatens their established teaching practices, while practitioners may fear being replaced by these technologies. To effectively address this resistance and successfully integrate GenAI, it is crucial to showcase how AI can augment and elevate existing approaches rather than replace them. By highlighting concrete examples of AI's benefits, such as providing personalized learning experiences through real-time individualized feedback,  educators can develop a more nuanced understanding of its potential and its ability to enhance their work. For dental practitioners, this might involve demonstrating how GenAI can help support their decision-making, freeing up time for more complex procedures and enhancing patient care \cite{Uribe2024, Preiksaitis2023, Islam2022, Li2023}.
    \item[Definition of roles and responsibilities] In implementing GenAI into the academic or clinical practice, it is essential to establish clear roles and responsibilities to ensure effective management and avoid task overlap. For example, if GenAI is to be implemented at a Dental School, it is crucial to establish a team responsible for managing the AI platform in both the clinical and educational programs, with specific roles such as data administrators, content developers, and ethics specialists \cite{Islam2022}. Establishing these clear roles and pipelines facilitates the practical implementation of AI in its intended use, and promotes compliance of the medical and administrative personnel.
    \item[Ethical considerations] Ethical considerations assume crucial importance when integrating  GenAI into the dental field, particularly in protecting the privacy of patients, practitioners, and students. The potential for unintended consequences is high when clinical data is used to train GenAI models, making it essential to ensure the anonymity and security of patient information at all stages. Clear policies must be established to mitigate the risks associated with GenAI, including governing the use of sensitive data, promoting transparency in AI decision-making processes, and safeguarding against biased outcomes. Furthermore, it is crucial to engage in open discussions about ethical  considerations and develop guidelines that balance the benefits of GenAI with the need to protect individual privacy and autonomy \cite{Gordon2024, Bannister2023, Saghiri2021, Mu2024, Preiksaitis2023}.
    
    On the other hand, there are still several concerns regarding the ethical and good practices behind using GenAI in biomedical research. For this, many journals and publishers have incorporated clear, written guidelines regarding how GenAI should be utilized and/or disclosed in research papers \cite{Ganjavi2024}. Furthermore, privacy concerns remain regarding the protection of sensitive patient data used to train GenAI models, as well as worries related to racial or other bias in the resulting generated data \cite{Chen2024,Gupta2023,Reddy2024}. On the other hand, the use of partially synthetic placebo groups can reduce the number of participants needed in clinical trial studies, leading to important bioethical advantages \cite{Azizi2021,Umer2024}. 

    \item[AI literacy] Educating students, academics, practitioners and researchers on ethically interacting with GenAI is essential. For example, offering courses or workshops that teach the users how to interpret and criticize the results of GenAI models can foster critical thinking in their clinical application and academic integrity. Additionally, providing examples of best practices in the use of AI in dentistry and highlighting cases where technology has improved diagnostic accuracy or treatment design can be beneficial \cite{Bannister2023, Preiksaitis2023, Islam2022, Li2023, Furfaro2024}.
    \item[Dependency on technology] Despite all the promising advances in AI, the over-reliance on GenAI in dental education and practice can have consequences, potentially suppressing the development of essential critical thinking skills among students and clinicians. A prime example of this is when professionals become too reliant on AI-driven models to make clinical decisions, which can lead to a lack of contextual understanding, ultimately compromising patient care. Moreover, this over-reliance may hinder the ability to adapt to novel or unexpected situations and diminish the capacity for independent problem-solving \cite{Gordon2024, Mu2024, Kim2023}. 
\end{description}

\section{Conclusion}
As discussed throughout this review, GenAI has the potential to revolutionize the field of dentistry by providing personalized and effective learning experiences for dental students, enhancing patient care and accelerating dental research. However, the successful implementation of GenAI in the dental field faces a series of challenges that must be addressed, including resistance to change, the definition of roles and responsibilities, ethical and privacy considerations, AI literacy, and the potential for over-reliance on technology. To overcome the challenges of deploying GenAI systems in dental environments, it is essential to educate instructors, practitioners, and researchers on AI-related topics, ensure the ethical use of sensitive patient data, and provide clear guidelines for developing and deploying GenAI models in dentistry. Additionally, educators and researchers must be aware of the limitations and potential biases of GenAI models and develop strategies to mitigate these risks. Ultimately, the successful integration of GenAI in the dental field will require a collaborative effort from educators, practitioners, researchers, and policymakers. By working together, we can harness the potential of GenAI to improve patient care, enhance dental education, and accelerate dental research, ultimately leading to better oral health outcomes for patients.

\section*{Funding}
This work was funded by the ANID National Doctoral Scholarship 21220200 (FV) and 21242208 (RG), and ANID Fondecyt Regular 1220804 (SA).

\section*{Declaration of generative AI and AI-assisted technologies in the writing process}

During the preparation of this work the authors used Llama 3 (Meta) and ChatGPT (OpenAI) in order to improve the language and readability of the paper. After using these tools, the authors reviewed and edited the content as needed and takes full responsibility for the content of the published article.

\bibliographystyle{elsarticle-num}

\bibliography{bibliography}

\end{document}